\documentclass[sigconf]{aamas}  
\usepackage{balance}  

\usepackage{booktabs}

\usepackage{bm}
\usepackage{subcaption}
\usepackage{mathtools}
\usepackage{color,soul}
\usepackage{pdfpages}
\usepackage{comment}
\usepackage{booktabs}
\usepackage{multirow}
\usepackage[normalem]{ulem} 

\setcopyright{ifaamas}  
\acmDOI{}  
\acmISBN{}  
\acmConference[AAMAS'19]{Proc.\@ of the 18th International Conference on Autonomous Agents and Multiagent Systems (AAMAS 2019)}{May 13--17, 2019}{Montreal, Canada}{N.~Agmon, M.~E.~Taylor, E.~Elkind, M.~Veloso (eds.)}  
\acmYear{2019}  
\copyrightyear{2019}  
\acmPrice{}  

\newif\ifcomments
\commentstrue

\ifcomments
\newcommand{\commentjb}[1]{\textcolor{blue}{(JB: #1)}}
\newcommand{\commentsj}[1]{\textcolor{red}{(SJ: #1)}}
\newcommand{\commentam}[1]{\textcolor{brown}{(AM: #1)}}
\newcommand{\commentmk}[1]{\textcolor{maginta}{(MK: #1)}}
\newcommand{\old}[1]{\textcolor{red}{\sout{#1}}}

\else
 \newcommand{\commentjb}[1]{}
 \newcommand{\commentsj}[1]{}
 \newcommand{\commentam}[1]{}
 \newcommand{\commentmk}[1]{}
 \newcommand{\old}[1]{}
 
\fi

\begin{document}

\title{Domain Authoring Assistant for Intelligent Virtual Agents}  


\author{Sepehr Janghorbani}
\affiliation{%
  \institution{Rutgers University, Disney Research}}
\email{sepehr.janghorbani@rutgers.edu}

\author{Ashutosh Modi}
\affiliation{%
  \institution{Disney Research}}
\email{ashutosh.modi@disneyresearch.com}

\author{Jakob Buhmann}
\affiliation{%
  \institution{Disney Research}}
\email{jakob.buhmann@disneyresearch.com}

\author{Mubbasir Kapadia}
\affiliation{%
  \institution{Rutgers University}}
\email{mubbasir.kapadia@rutgers.edu}

\begin{abstract}  
Developing intelligent virtual characters has attracted a lot of attention in the recent years. The process of creating such characters often involves a team of creative authors who describe different aspects of the characters in natural language, and planning experts that translate this description into a planning domain. This can be quite challenging as the team of creative authors should diligently define every aspect of the character especially if it contains complex human-like behavior. Also a team of engineers has to manually translate the natural language description of a character's personality into the  planning domain knowledge. This can be extremely time and resource demanding and can be an obstacle to author's creativity. The goal of this paper is to introduce an authoring assistant tool to automate the process of domain generation from natural language description of virtual characters, thus bridging between the creative authoring team and the planning domain experts. Moreover, the proposed tool also identifies possible missing information in the domain description and iteratively makes suggestions to the author.
\end{abstract}

%


\maketitle


\section{Introduction}\label{sec:intro}

Intelligent Virtual Agents (IVA) technology has been applied in many fields such as education \cite{kim2008virtual} or entertainment, where the agents are used for creating virtual user experiences or for digital storytelling \cite{theune2003virtual}.  
Interactive virtual agent design has been the focus of research for the past two decades~\cite{fong2003survey}. In an effort to give agents deliberative capabilities (a key requirement for social interactions with humans and other agents), a common approach is to use a planner to model the agent's decision making process \cite{youssef2015towards,ribeiro2014thalamus,matsuyama2016socially}. Planning architectures are 
well suited for this problem, since the world of these agents can be 
modeled as a set of discrete objects, and naturally maps to a logic-based planning domain language.
For instance, the world can be modeled by a set of smart-objects \cite{kallmann1999modeling}, each 
having a set of discrete states and set of affordances \cite{gibson2014ecological}, where the latter represents the advertised actions of that object. 

\begin{figure*}[t!]
    \centering
    \includegraphics[width=0.8\textwidth]{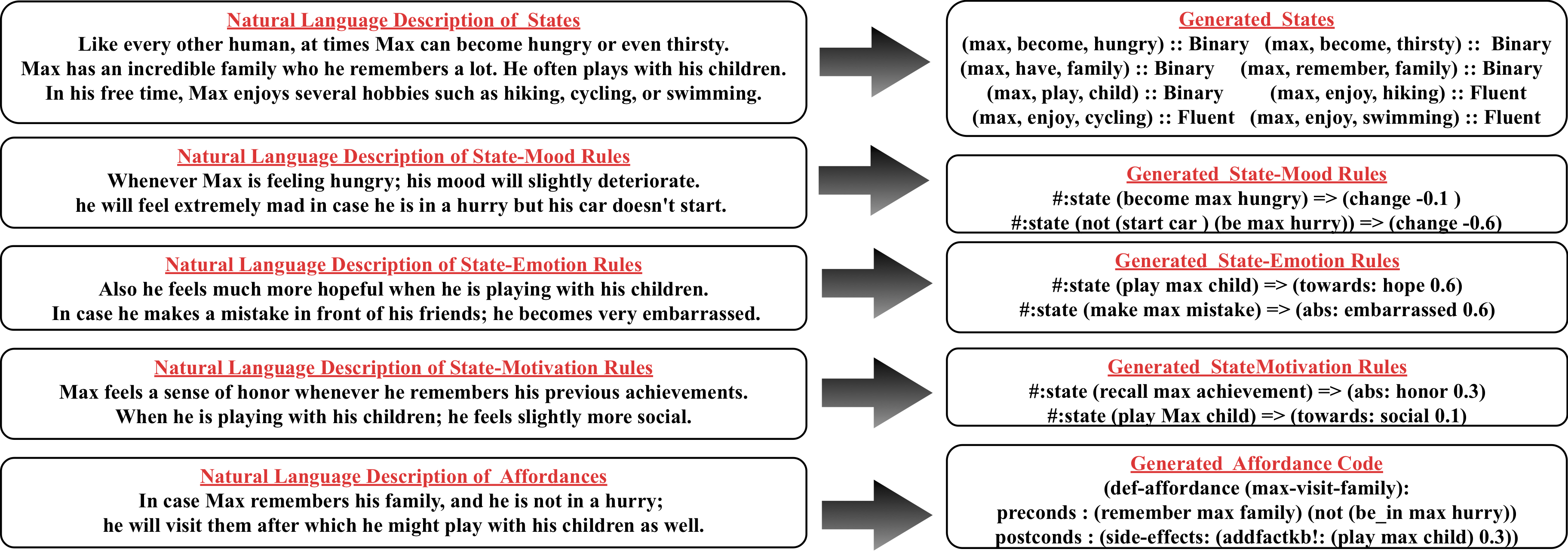}
     \caption{Input (left) and output (right) of the proposed model for a small sample domain}
     \label{Fig:exampleTranslation}
\end{figure*}

Intelligent agents, especially those designed to exhibit plausible social interactions with human users~\cite{cassell2000embodied}, must exhibit deliberative capabilities, express emotion and personality traits~\cite{gebhard2005alma}, and have mental models of their application domain (e.g., interactive games~\cite{harrington2016video}, or storytelling~\cite{theune2003virtual}). In order to meet these requirements, creative authors 
must carefully design aspects of the character's personality and emotion profile, its motivations and its representation of the virtual world it resides in. Creative authors typically use natural language descriptions to design these characters. For example, for describing the emotional profile of Max, (the example character in this paper), a creative would say: ``Max becomes slightly angry in case he sees his favorite sports team lose''.  These natural language descriptions of the agent's behavior need to be translated to a machine-readable planning language, i.e.\ the domain knowledge of a planner. This translation is manually performed by a team of domain experts, however it is time consuming and resource intensive, and requires frequent interactions between the creative team and the domain experts.

This paper proposes an automated process for translating the natural language description of an agent's behavior to the planning domain knowledge. Authors when describing agent's behavior are typically constrained by domain restrictions, however, the proposed system aims to assist the creative authors in defining, more freely, the behavior of an intelligent character in natural language. 
The availability of such a tool would not only expand the ease of authoring of the IVAs, but would also promote the applicability of such agents in general, since one of the main bottlenecks in creating IVAs lies in the generation of the agent's domain and its affective states.

The research problem addressed in this paper is inherently challenging due to a number of reasons:
(1) non-triviality of the task of automatically understanding natural language,
(2) generating an executable planning domain despite the variability in the writing style,
and (3) giving the authors as much freedom as possible while still directing them to specify the domain-related material.

Supervised machine learning approaches have been proven to be useful for a similar task of semantic parsing \citep{SemanticParsing2018}. However, such statistical models cannot be trained for the task at hand due to the lack of sufficient amount of labeled data. In this paper we take an alternate approach for parsing the sentences using dependency graphs \cite{spacy2}. Dependency graphs explicitly describe the syntactic relations between different words of the sentence, and hence are useful in extracting information relevant to our task.


While researchers have worked on similar areas of planning domain acquisition, such as robot teaching~\cite{lindsay2017framer,kollar2013learning} as well as in the task of story comprehension \cite{diakidoy2015star,hayton2017storyframer}, to the best of our knowledge, we are among the first to address the problem of domain acquisition for developing character-based intelligent virtual agents.

\section{Related Work}\label{sec:related}

It has been shown that the natural language description of a planning domain 
is as expressive as the formal planning domain definition itself \cite{branavan2012learning}. 
Several researchers have investigated the problem of planning domain generation from natural language text.
For example, a highly investigated field of study is automatic planning domain acquisition from natural language instructions directed at robots ~\cite{lindsay2017framer,kollar2013learning,perera2014task,gemignani2015teaching}.
These instructions can be translated to the planning domain by taking a rule-based parsing approach similar to ours \cite{lindsay2017framer}.  
\citeauthor{perera2014task} \cite{perera2014task} address the same problem but only parse simple and direct instructions such as ``go to'' and ``deliver'' into single actions, and their model does not generalize to a complete domain. Similarly, \citeauthor{Pomarlan-2018} \cite{Pomarlan-2018} construct an executable robot program using the Cognitive Robot Abstract Machine (CRAM) language, produced by parsing natural language instructions.

The model of ~\citeauthor{yordanova2017texttohbm} \cite{yordanova2017texttohbm} uses textual instructions for human activities to learn the actions in the planning domain as well as their pre/post-conditions (cf.\ Section~\ref{sec:agentmodel}).
Their method builds an ontology of the domain and then optimizes the model based on sensory data. This approach is suited for the generation of a highly constrained robot planning domain and is less helpful for the highly unconstrained task of intelligent character development.

The generation of action pre-conditions from natural language instructions is addressed by \citeauthor{bindiganavale2000dynamically} \cite{bindiganavale2000dynamically}.
Their system works by extracting simple subject-predicate relations in the sentence using an XTAG grammar system. However, the system is suitable for development of simulation environments and not for the task of character development since it gives reasonable performance only in case of simple and direct instructions and does not cover the complete domain.
Similarly \citeauthor{goldwasser2014learning} \cite{goldwasser2014learning} train a model to learn a target representation of states and domain rules from a textual description of a simulated environment, e.g., a computer game. They train the parameters of their model with a small number of training data points. Additionally, their model also has a feedback mechanism which generates positive/negative training labels after performing a predicted action. These approaches are suited for development of simulated environments, but for intelligent character development, neither simple instructions nor simulating the agent for feedback generation can be used. 

The model proposed in \cite{sil2011extracting} combines the idea of planning domain acquisition with common sense knowledge information, by using semantic role labeling and large-corpus statistics. In our work, we also utilize common sense knowledge, however in a supportive feedback mechanism directed at the authors. The proposed model \cite{sil2011extracting} tries to learn STRIPS (a general-purpose planning domain representation), from web-based data by utilizing common sense knowledge base queries to infer some of the implicit pre/post-conditions. 

Another line of work related to intelligent character domain generation is the task of story comprehension \cite{diakidoy2015star,sanghrajka2018computer,hayton2017storyframer}.
Given a natural language narrative of the story, the goal of this line of research is to understand the key plot of the story as well as the set of events.
STAR \cite{diakidoy2015star} is an automated story comprehension tool for extracting the key events of a story and doing inference based on the order of the events. It also introduces the concept of a world knowledge in stories, i.e.\ a set of universal rules governing a wide set of different stories.

\citeauthor{sanghrajka2018computer} \cite{sanghrajka2018computer} extend STAR by applying a logical reasoning system to detect inconsistencies in the story plot, making inferences about the implicit knowledge in the story. 
Both systems relate to the concept of story comprehension based on inference from a knowledge-base, similar to the ``Common Sense Module'' in our system.  

Our proposed approach is conceptually similar to StoryFramer \cite{hayton2017storyframer}, which infers the narrative planning domain based on a plot described in natural language. The authors apply linguistic rules on parsed text to connect events with predefined pre/post-conditions. However, our approach differs from StoryFramer as follows: (1) Our system is designed to author the entire domain of an intelligent virtual agent, which includes its inner state, affordances, personality, mood, emotion, and motivations, and a representation of world state. This domain significantly builds upon the narrative domain utilized in StoryFramer. (2) Our system supports more
complex, compound sentence constructs, which are needed to author the domain described above. (3) Our system supports the automatic extraction of affordance pre- and post-conditions directly from natural language descriptions, thus facilitating generalizability and scalability to completely novel planning domains. 



\begin{figure}[t]
\includegraphics[width=7cm]{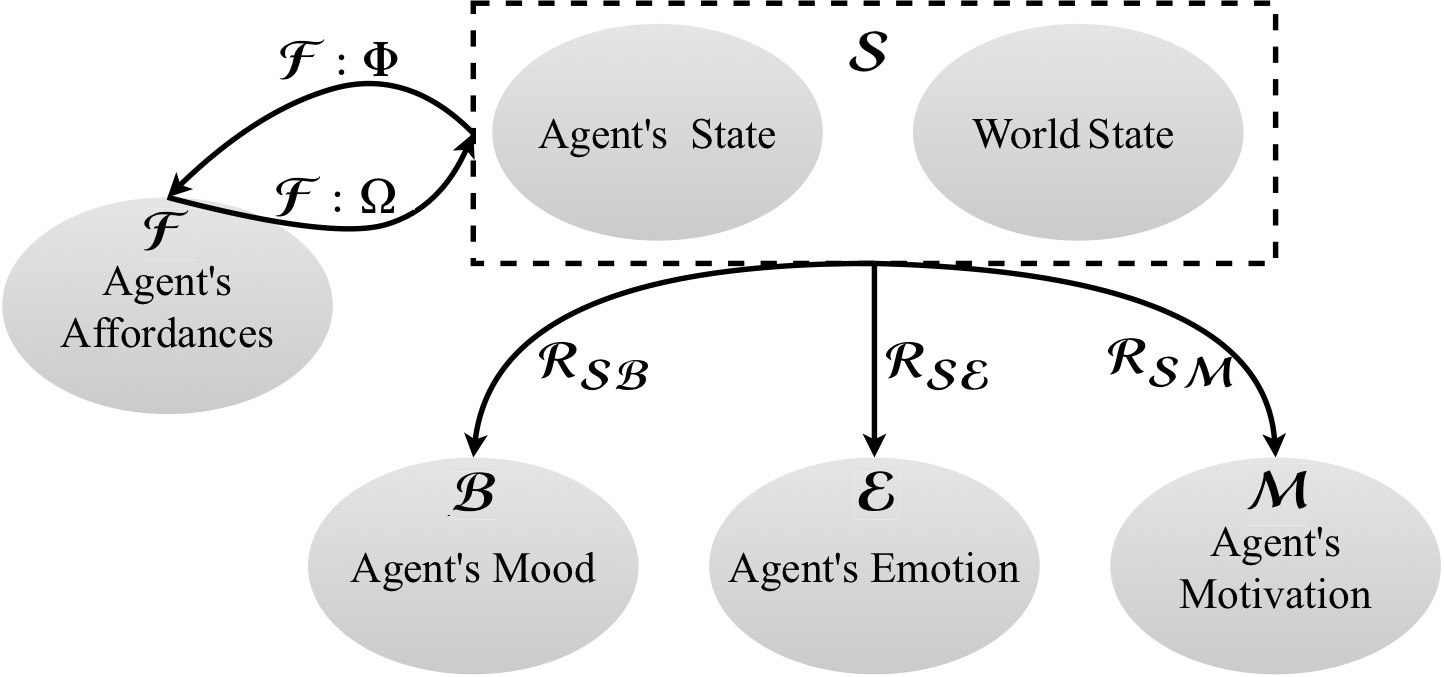}
\caption{Schematic diagram showing the agent architecture proposed in this paper. It shows the interaction between the different components of the model as described in Section~\ref{sec:problemDef}.
}
\label{Fig:emotionModel}
\end{figure}

\section{Agent Model}
\label{sec:agentmodel}

In this work, the agent architecture and its world is built around the concept of \textit{smart-objects} \cite{kallmann1999modeling} where the agent itself is also an active smart-object, and the world describes the set of all smart-objects.

Each smart-object is defined by a set of \textit{states}, representing a set of facts about the smart-object, and \textit{affordances} \cite{gibson2014ecological}, which reflect the set of offered capabilities on how the smart-object can change its state or the state of other smart objects. Each affordance is defined by a set of pre-conditions and a set of post-conditions.
While the state changes are captured in the post-condition of the affordance, its pre-conditions represent the logical conditions on the states which have to be satisfied so that the affordances can be executed.

In our approach, smart-objects with all their states and affordances, including the logic of pre/post-conditions, are inferred directly from a natural language description. Figure~\ref{Fig:exampleTranslation} shows the input and the generated output for a small example domain. Notice that the post-condition of the affordance is satisfied in a non-deterministic way.

Since the agent architecture is designed for an intelligent character, the agent's state contains not only information about its physical state and the states of the environment, but also the agent's mood, its emotions, and its motivations (Figure~\ref{Fig:emotionModel}). While emotions represent short term variables of affect, the mood is a long term concept shaping agent's behavior. The motivation states are used to build the objective function that is used by the planner - in our case a heuristic based search algorithm - that generates plans bringing the agent closer to a set of target motivation values. All three sets of variables are influenced by state-dependent rules, as symbolized by the arrows in Figure~\ref{Fig:emotionModel}.

Although our implementation focuses on this particular agent architecture, the methods developed in this paper are fairly general and can be extended to other agent architectures~\cite{gratch2004domain,pecune2016socrates}, or planning domains (e.g., PDDL) as well. For instance, when the behavior should be influenced by social rules, when the utility of the agent's goal is not encoded with motivations but another objective, or when the affective state are modeled with other mechanisms (or not at all), natural language can be used to build the domain and facilitate the authoring process. Future research will focus on making the approach more general to a broader class of agent architectures in this field.

\begin{figure}[t]
    \centering
    \includegraphics[width=8cm]{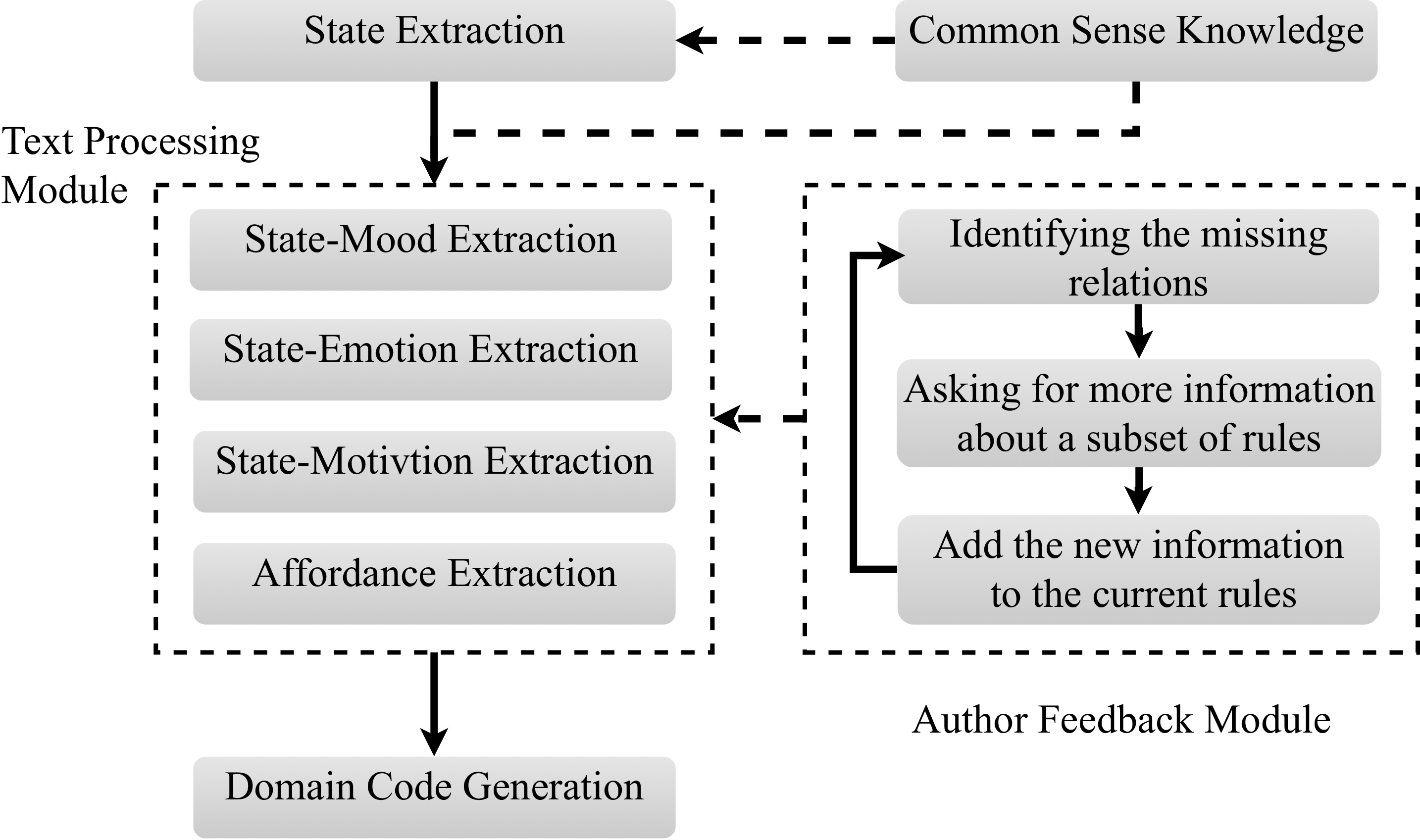}
     \caption{Pipeline for going from natural language description to domain code, as described in Section~\ref{sec:methods}. Solid arrows represent actual data transfer while dashed arrows indicate giving suggestions to the author.}
     \label{Fig:method1}
\end{figure}


\subsection{Problem Domain Definition}
\label{sec:problemDef}

Formally, the agent is defined by a tuple $\boldsymbol{\Sigma}= \boldsymbol{ \langle \mathcal{S}}, \boldsymbol{\mathcal{B}}, \boldsymbol{\mathcal{E}}, \boldsymbol{\mathcal{M}}, \boldsymbol{\mathcal{F} \rangle}$,
with $\boldsymbol{\mathcal{S}}$ representing the state space,
$\boldsymbol{\mathcal{B}} $ the agent's mood, $\boldsymbol{\mathcal{E}}$ the agent's set of emotions, $\boldsymbol{\mathcal{M}} $ its motivations, 
and $\boldsymbol{\mathcal{F}}$ the set of affordances.
Each affordance $\boldsymbol{f} \in \boldsymbol{\mathcal{F}} : \boldsymbol{f}=\boldsymbol{\langle \mathcal{O}}, \boldsymbol{\Phi}, \boldsymbol{\Omega}, \boldsymbol{\Lambda \rangle}$ is itself a tuple of the affordance owner $\mathcal{O}$, a set of pre-conditions $\boldsymbol{\Phi}$, a set of post-conditions $\boldsymbol{\Omega}$, and side effects $\boldsymbol{\Lambda}$.

\begin{figure*}[t]
    \centering
    \includegraphics[width=0.9\textwidth]{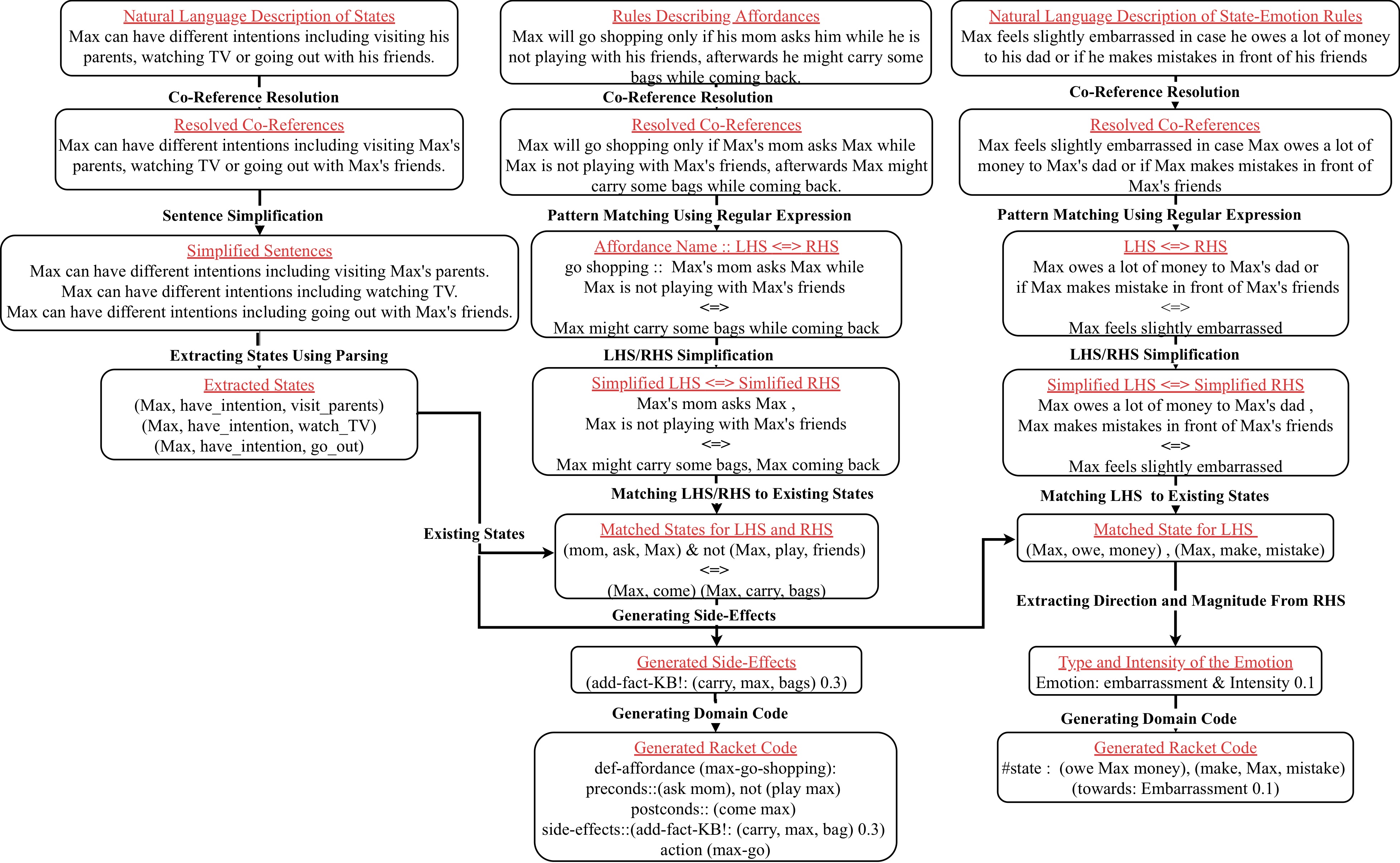}
    \caption{Detailed pipeline with examples for each submodule.}\label{Fig:method2}
\end{figure*}

\begin{itemize}
\item

$\boldsymbol{\mathcal{S}}$ represents the set of agent's states and the state of the world, see Figure~\ref{Fig:emotionModel}. The model differentiates between binary states (e.g.\ whether the agent is asleep or not) and $n$-ary states (e.g.\ 
agent's position), also referred to as \textit{fluents}. 
 
\item
$\boldsymbol{\mathcal{B}} $ represents the agent's mood.
It is a continuous variable in the range $[-1,1]$, where $-1$ represents a negative mood, $0$ a casual mood, and $1$ a positive mood.

\item
$\boldsymbol{\mathcal{E}}$  represents the current emotion the agent is experiencing. The emotions are modeled using the Pleasure, Arousal, Dominance (PAD) formalism \cite{mehrabian1996pleasure} which associates to each emotion a position in the 3-dimensional PAD space \cite{gebhard2005alma}.

\item
$\boldsymbol{\mathcal{M}} $ represents motivations which  drive agent's decision making system. The motivations are modeled using Reiss Motivational Profiles (RMP) \cite{reiss1998toward} covering the full range of motivations into a set of 16 
motivational factors, each factor being a number in the range $[0,1]$. 
Examples of such motivations include ``honor'', ``order'' and ``family-relationships'', etc.

\item
$\boldsymbol{\mathcal{F}}$ is the agent's set of affordances.
These represent the set of actions that can be performed to change the states of one or more smart-objects.
An affordance can be applied when a logical conjunction of states (i.e. pre-conditions $\boldsymbol{\Phi}$) is satisfied, and upon successful execution it changes states according to its post-conditions $\boldsymbol{\Omega}$. Additionally, the affordances in this particular agent architecture may contain a set of so-called side-effects, $\boldsymbol{\Lambda}$, which allow the affordance to change states in a non-deterministic way.
\end{itemize}

The agent's architecture contains several mechanisms to model the changes in the agent's affective states (emotions, mood, and motivations). These mechanisms are rule based functions
 $\boldsymbol{\mathcal{R(S,A)}} : \boldsymbol{\mathcal{S_{s}}} \to \boldsymbol{\mathcal{A} }$ from states to affects, where $\boldsymbol{\mathcal{S_{s}}}$ is a conjunctive normal form (CNF) on states $\boldsymbol{\mathcal{S}}$ and $\boldsymbol{\mathcal{A} } \in \{ \boldsymbol{\mathcal{B}},\boldsymbol{\mathcal{E}}$, $\boldsymbol{\mathcal{M}} \}$.

In this paper, we interchangeably use the terms pre-condition and left hand side (LHS) and similarly the terms right hand side (RHS) and post-condition. 
The 
rule-based functions are described below: \\ 

\noindent\textbf{State-Emotion Rules} $\boldsymbol{ \mathcal{R(S,E)}}$ address the changes of the emotions, the position in PAD space, based on the current state. 
The RHS reflects a strength and direction of the shift of the PAD values towards an emotion (e.g increase 0.2). See the right column of Figure~\ref{Fig:method2}. \\

\noindent\textbf{State-Mood Rules} $\boldsymbol{\mathcal{R(S,B)}}$ describe how mood changes given the current states. While the LHS is a conjunctive normal form on the state space, the RHS specifies the direction and step size of the mood change. \\

\noindent\textbf{State-Motivation Rules} $\boldsymbol{ \mathcal{R(S,M)}}$ represent the mapping from states to the motivations. The RHS either specifies the direction and the strength of change towards any of the motivations or it directly sets the motivations to a specific value.

This paper proposes a model for generating $\boldsymbol{\langle \mathcal{S}}, \boldsymbol{\mathcal{F}}, \boldsymbol{ \mathcal{R_{SB}}}, \boldsymbol{\mathcal{R_{SE}}}, \\ \boldsymbol{\mathcal{R_{SM}} \rangle}$, given a natural language description of the domain provided by a creative author. 

We assume that the set of possible affective states ($\boldsymbol{ \mathcal{B}}, \boldsymbol{\mathcal{E}}, \boldsymbol{\mathcal{M}}$) are predefined and known by the authors.
Our proposed system must guarantee that it finds a consistent set of states and is able to use it for generating the logical components the planning system relies on. This effort involves multiple challenges as discussed in the following sections.


\section{Domain Generation from Text}\label{sec:methods}

For generating domain code from natural language descriptions we propose a modular system, as shown in Figure~\ref{Fig:method1}.
The first step in the system pipeline is the
\textit{State Extraction Module}, which is used to identify and extract the state-related information from each sentence, and subsequently used to construct the states of the smart-objects as well as the states of the world $\boldsymbol{\mathcal{S}}$. Next, the \emph{Text Processing Module} extracts the affordances as well as the state-affect rules using similar mechanisms (see  Section~\ref{sec:stateExtraction}). 
Lastly, the \textit{Domain Code Generation Module} generates a computer readable planning language from states, affordances and state-affect rules.

Figure~\ref{Fig:method2} exemplifies the entire pipeline. Since the mechanisms used in all state-affect extraction modules are very similar, only the state-emotion extraction is shown. 

For the agent architecture in this paper, Racket \cite{plt-tr1} was used as the target language, however, generalization to other general purpose planning languages (e.g.\ PDDL) is straightforward \cite{mcdermott98pddl}. 

In addition to the processing steps described above, we introduce two more modules that significantly help in generating domain from natural language: (1) a \emph{Common Sense Knowledge Module} and (2) an \emph{Author Feedback Module}. 
The common sense knowledge module (cf.\ Section~\ref{sec:commonsense} ) facilitates the authoring by suggesting new information about affordances or new rules regarding affective states. For these suggestions a common sense knowledge-base is queried, i.e., ConceptNet \cite{speer2012representing}. Besides finding missing information, this module should also promote the creativity of the authors by providing new and relevant information about the already authored objects. 
Furthermore, in case of an unclear natural language description, the author feedback module (cf.\ Section~\ref{sec:authorFeedback}) can inform the creative author of possibly missing information about the domain, such as incomplete affordances or possible state dependencies.

\subsection{State Extraction}\label{sec:stateExtraction}

\begin{figure}[t!]
    \centering
    \includegraphics[width=8cm]{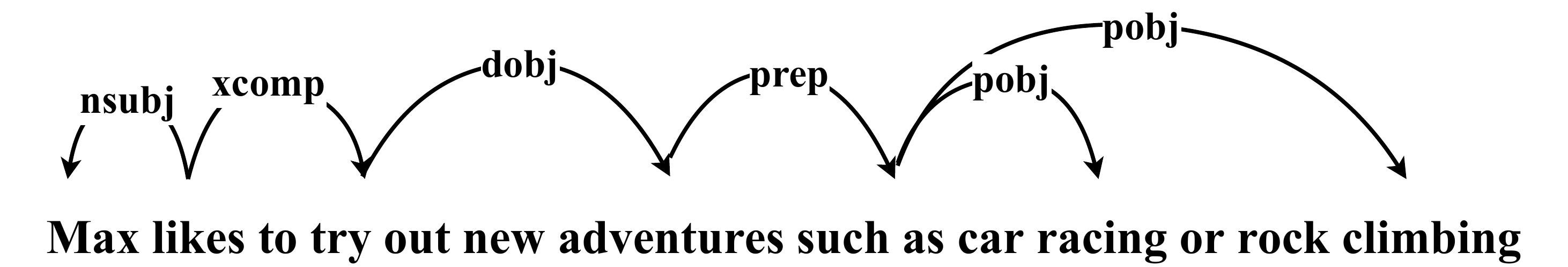}
    \caption{Dependency parse graph for a given sentence from which underlying states: (Max, try\_out, racing) and (Max, try\_out, climbing) are extracted. The abbreviations used in the figure are as follows: (nsubj: noun subject, xcomp: open clausal complement, dobj: direct object, prep: prepositional modifier, pobj: object of preposition). 
    }\label{Fig:dependencyParsing}
\end{figure}

Natural language text can be ambiguous w.r.t.\ entities mentioned in the text. To resolve these ambiguities about the entity references (e.g., entity referred by a pronoun) we perform co-reference resolution.
For example, in the sentence ``Max brings the book and then \textit{he} reads it.'', from the point of computational language processing, it is ambiguous whether ``he'' refers to Max or the book. The co-reference module helps to resolve it. The normalized sentence after the co-reference resolution is ``Max brings the book and then \textit{Max} reads it.'. 
The co-reference normalized sentences are further simplified using rules based on lexical substitution and syntactic reduction techniques \cite{saggion2017automatic}. These rules reduce a complex sentence into multiple simpler sentences each with a single main verb. In order to extract state related information, each of the simple sentences is then parsed using a dependency parser \cite{spacy-parser}. Dependency parser processes a sentence to produce a dependency graph which gives syntactic relations (e.g., subj, obj, etc.) between the words in the sentence.  Figure~\ref{Fig:dependencyParsing} shows an example of a dependency graph obtained for an example sentence. 
Linguistic simplification rules are then applied on the dependency graph to extract information used for constructing the states. These linguistic rules were designed based on several sample writing styles provided by professional 
creative authors, and we tried to keep the rules as general as possible. We describe these rules in the following paragraphs. 

Before the text is simplified based on linguistic rules, types of states mentioned in the sentences are identified, i.e.\ either fluent (``$n$-ary'') or ``binary''. This identification is done by checking the presence of certain keywords in the sentence that are generally used to describe fluent variables. Examples of keywords used are ``including, such as, consist of'', etc. The sentences are then processed with different sets of  linguistic rules depending on the state type. 

Since a sentence corresponding to a fluent state has one of the mentioned keywords, it typically follows a specific syntactic structure
Generally, the mentioned 
keywords in sentences containing a fluent, syntactically occur as the prepositional modifier and have prepositional complement (pcomp) dependency relationship with the next word, which is usually a content-rich word carrying important information. For such sentences, words corresponding to the syntactic relations such as prepositional modifier, prepositional complement and their object are extracted. An example of each of these relations as well as the extracted states is provided in Table~\ref{Tab:langRules}. 
The extracted state is constructed with a triplet: subject, verb (+ object) and prepositional complement (+ object). 
Contrary to sentences with fluents, sentences with binary states can have more diverse syntactic patterns. This requires extraction of different and more diverse set of syntactic relations. 
For such sentences, words corresponding to the relations such as prepositional modifier, adjectival complement, open clausal complement alongside their objects (see Table~\ref{Tab:langRules}) are extracted.  

In Table~\ref{Tab:langRules}, the first two rows show examples of sentences with a fluent while the rest are sentences containing binary states. The linguistic rules shown here serve as illustrative examples, and can easily be extended to be more flexible, or  accommodate a wider and more diverse range of writing styles.

\begin{table*}[t!]
\small 
 \centering
  \caption{Linguistic Rules used in the system. Bold words are the specified relation. solid underline is the verb, dashed is the subject and dotted is the main verb's object. }
    \begin{tabular}{p{0.27\linewidth} p{0.40\linewidth} p{0.25\linewidth} }
    \toprule
    \textbf{Relation Name} & \textbf{Example Sentence} &    \textbf{Extracted State} \\ \toprule

  Prepositional Modifier and its object
  &   \dashuline{Max} can \ul{go} to different \dotuline{places} such \textbf{as} \textbf{restaurants} and \textbf{parks} &     (Max, go, restaurant) (Max, go, park)\\ \midrule
  
    Prepositional Complement and its object  &  
    \dashuline{Max} can \ul{engage} in different \dotuline{activities} including \textbf{riding} a \textbf{horse}. & (Max, engage\_in, ride\_horse)\\ \midrule
    
    Adjectival Complement and its object &  \dashuline{Max} can \ul{be} \textbf{aware} of his \textbf{surroundings}. & (Max, be\_aware, surrounding) \\ \midrule
    
    Preposition and its object & \dashuline{Max} can \ul{stand} \textbf{at} the \textbf{bus station}. & (Max, stand, station)\\ \midrule
    
  Open Clausal Complement and its object & \dashuline{Max} would \ul{like} to \textbf{drink} some juice &  (Max, drink, juice) \\
    
    \bottomrule \\

    \end{tabular}
    \label{Tab:langRules}                           

\end{table*}

\subsection{Affordance Extraction}\label{sec:affordanceExtraction}
To identify the RHS and LHS of the affordance descriptions, the co-reference normalized text is processed with a pattern matching algorithm. These diverse sets of patterns have also been defined based on a creative author's recommendation for natural writing style and provide flexibility to the authors in describing affordances. For example, in the sentence ``Max goes to the library \textit{only if} he has an exam \textit{after which} he feels more knowledgeable.'', the keywords \textit{only if} and \textit{after which} are separating different parts of the affordance. The part before the first keyword is used for obtaining the name of the affordance while the next two parts constitute the set of pre-conditions and post-conditions respectively. Similarly, other patterns are also defined to obtain the pre/post-conditions and to generalize over different language formulations. These patterns can be easily extended or customized.

After extracting the LHS and RHS using the pattern matcher and subsequently simplifying LHS/RHS, the rule-based parsing approaches from Section~\ref{sec:stateExtraction} are used to extract the states. Next, the extracted state names are mapped to the already existing states obtained from the State Extraction Module based on semantic similarity. 
We use averaged word2vec word embeddings \cite{mikolov2013efficient} to map the names to a vector (embedding) space where cosine similarity is used to match the names. An embedding is a vector representation capturing the syntactic and semantic properties of the word. Distributional hypothesis \cite{harris1954distributional} states that words occurring in similar context have similar meanings. Typically, these embeddings are learned from co-occurrence 
statistics of words in large text corpora. Consequently, semantically similar words have embeddings which are closer to each other in the vector space \cite{mikolov2013distributed}.

To be robust against noise, generic verbs (e.g.\ light verbs such as `be') are filtered out before mapping to the embedding space, if they do not have a context-specific meaning. 
Finally, each extracted post-condition is checked for probabilistic connotation, by searching for keywords corresponding to a notion of uncertainty, such as ``probably, possibly, definitely''. Each  keyword is assigned a different predefined probability, which quantifies the non-deterministic nature of the post-condition.
 
\subsection{State-Affect Extraction }
\label{sec:affectExtraction} 
State-mood rules $\boldsymbol{\mathcal{R(S,B)}}$, state-emotion rules $\boldsymbol{\mathcal{R(S,E)}}$, and state-motivation rules $\boldsymbol{\mathcal{R(S,M)}}$ are extracted in a similar manner as the Affordance Extraction Module with a few minor differences. 
In the first step, pattern matching is used but with a different set of patterns since the natural sentences in which authors describe states of affect are commonly different than affordance descriptions. For instance, in the sentence ``Max will get extremely angry \textit{whenever} he fails his exams.'', the part before \textit{whenever} is describing the affective change while the other part describes the state (LHS of the rule).
The next step of state extraction and matching is the same for the LHS. For the RHS, 
the direction and magnitude of the change is obtained by looking for certain adverbs. 
For instance, 
adverbs such as `extremely' or `very much' represent a high degree of change while others such as `moderately' stand for moderate degree of change.

\subsection{Author Feedback}
\label{sec:authorFeedback}
Sometimes the authors might forget to specify the interactions between states of the agent and its states of affect, or forget to specify the pre/post-conditions of an  affordance. The author feedback module attempts to identify such 
cases. Subsequently, the module comes up with suggestions for the authors. 

The module considers the set of all the possible missing rules which could be constructed. In order to show only the relevant suggestions to the author, a subset of the rules are selected. This is done by calculating the similarity between LHS and RHS of the rule and if the score is above a predefined threshold, it is added to the subset of the rules to be presented. The intuition being that if the LHS and RHS of the rule are similar enough, the rule is probably worth reviewing. For example, the state ``eating'' and the emotion ``hunger'' are related, which may be described by a rule missed by the author.

For the affordances with few pre/post-conditions, the module asks for additional clarification information.
One challenge here is to set the similarity threshold in such a way that it does not filter out too many states while also not suggesting too many irrelevant rules either. For our system, the threshold was determined empirically by experimenting with the system.

\subsection{Common Sense Knowledge Suggestion}\label{sec:commonsense}
ConceptNet \cite{speer2012representing} is a crowd-sourced online knowledge base of common sense knowledge containing over 8 million entities and 21 million relations. 
It perfectly suits our purpose since it contains the types of relations that can be used for defining affordances and states. This is in contrast to other knowledge bases like Yago \cite{suchanek2007yago} which contain only factual information about the world. The entries and relations in ConceptNet are used in three main ways to assist the authors: (1) For each of the smart-object, the authors also specify a type. These types are used to query the information from ConceptNet relevant to the edge `IsCapableOf', thereby suggesting to the author possible capabilities or states that are tied to this smart-object;  (2) relations such as `Causes', `Entails', `HasFirstSubevent', `HasLastSubevent, `CreatedBy', `HasPrerequisite' are used to suggest pre/post-conditions for the affordances by querying the affordance name; (3) `CausesDesire' and `MotivatedByGoal' are used to suggest states which can cause an emotional/motivational response from the agent. These are only suggested if the similarity is above a certain threshold (determined empirically). 
\begin{figure*}[t!]
    \centering
    \includegraphics[width=0.8\textwidth]{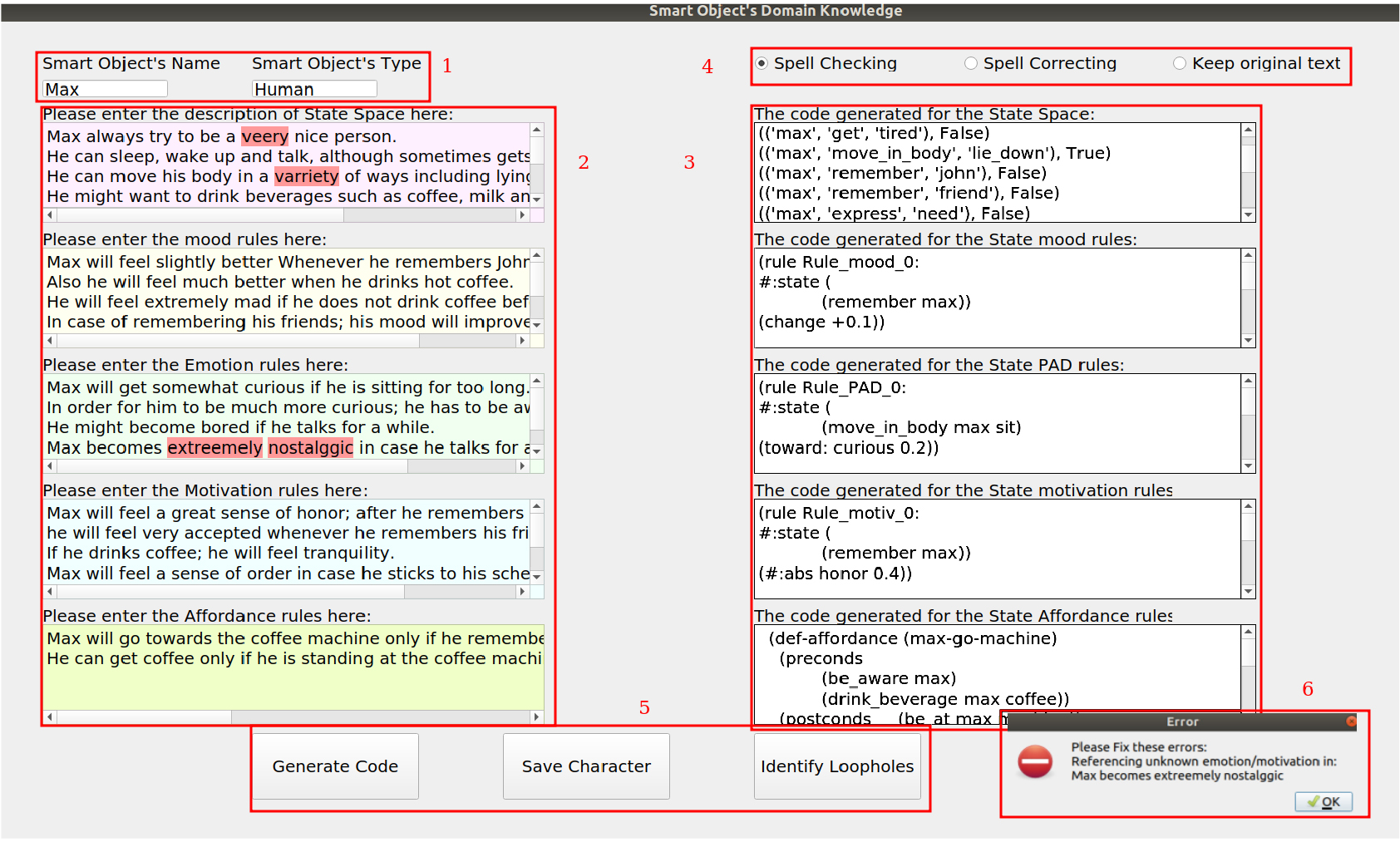}
    \caption{GUI tool for the system: (1) Specification of the character's name and type. (2) Input panels for natural language description of states and rules. (3) Generated output code from natural language description in (2). (4) Toggle for turning spell checking and correction on or off (see highlighted words in the input panels). (5) Various control buttons during authoring, and (6) pop-up of potential errors. 
    }\label{Fig:GUI2}
\end{figure*}

Since ConceptNet is crowdsourced, some entries are noisy, for this reason we only choose the more credible entries based on the number of contributors. This is controlled with the WEIGHT parameter. We set it to be at least 1. Table~\ref{tab:table-commonsense} provides some examples of suggestions provided by the proposed Common Sense Knowledge Module. The author is free to accept these suggestions to automatically expand their domain knowledge, or disregard them. As shown in Table~\ref{tab:table-commonsense}, suggestions like ``Max is a type of Dog, can it learn to do tricks?'' can give recommendations to the authors  regarding new ideas about aspects of the character previously not investigated.

\begin{table}[ht] 
 \centering
 \caption{Example suggestions provided to the author by the Common Sense Knowledge Module. }
    \begin{tabular}{p{0.9\linewidth} } \toprule

    Since ``Rio'' is a ``Bird'', can it ``prepare nest''?\\ \midrule 
Since ``Max'' is a type of ``Dog'', does it ``guide a blind person''?
 \\ \midrule 
Is ``fatten'', a post-condition of ``feed''?
\\ \midrule 
Is ``have guitar in hands'', a cause of ``play guitar''? \\ \midrule
Since ``Max'' is a type of ``Dog'', can it ``learn to do tricks'' ? \\
\bottomrule \\
    \end{tabular}
    \label{tab:table-commonsense}                            
\end{table}

\section{Evaluation}
\label{sec:evaluation}

To help the creative authors use our system as easily and seamlessly as possible, we have developed a simple GUI tool for our application, as shown in  Figure~\ref{Fig:GUI2}. 
This GUI is developed using PyQt5 \cite{summerfield2007rapid}. Spacy 2.0~\cite{spacy2} is used for co-reference resolution and dependency parsing due to its good performance, robustness and ease of use. The co-reference resolver was further improved to resolve possessive pronouns as well. Pre-trained word embedding (word2vec) are obtained using Gensim \cite{rehurek_lrec}, and the PyEnchant library\footnote{https://pypi.org/project/pyenchant/} is used for spelling mistake detection and correction.

The task of character development for intelligent virtual agents is a rather new field of study and to our knowledge, there is no standard benchmark or baseline for comparative evaluation of systems. Some authoring tools  such as STAR\cite{diakidoy2015star} are intended for story comprehension but they have an entirely different objective of identifying the main plots of the story, while in our case the goal is to extract the personality traits of an agent.
It is also not clear what should be the most appropriate evaluation metrics. Furthermore bringing the actual character to life needs running a planning mechanism which might not be available in many cases.

Nevertheless, as described in the next sections, for evaluating our tool, we quantitatively compare the output of our system with a ``gold standard dataset''. We also evaluate our system qualitatively by conducting a user study to assess the system's applicability.

\subsection{Quantitative Evaluation}
To quantitatively evaluate our system, we developed a ``gold standard dataset'' to be used for comparison.
The dataset was developed by a team of professional writers with diverse writing styles. Manual annotation of the natural language sentences was done by a team of domain experts 
to identify the corresponding domain knowledge. The dataset contains $26$ states, $23$ affordances and $6$ state affect rules. Each affordance on average has $1-2$ pre/post-conditions. By feeding the natural language component of the dataset into the model we generate the corresponding knowledge, which is then compared to the ground truth knowledge. \\

\begin{table*}[t]                           
 \centering
 \caption{User study results for (a) Tool-specific questionnaire, and (b) System Usability Study (SUS) questionnaire.}
 \small 
    \begin{tabular}{r p{9 cm} p{1.2cm} p{1cm} p{1cm} p{1cm} p{1cm} }
     \toprule 
    & \textbf{Question} & \textbf{Strongly Disagree} &    \textbf{Disagree}  &   \textbf{Neutral}  &   \textbf{Agree}  &   \textbf{Strongly Agree}  \\ 
    \toprule 
    \multirow{8}{*}{\rotatebox{90}{\textbf{Questions about tool}}}
    & I the tool useful for assisting in the authoring experience.     &   0.0\%    &   4.8\%  &   33.5\%  &   \textbf{47.6\%} &   14.3\%    
    \\ 
    
    & It was easy to learn the tool. &   9.5\% &   \textbf{33.3\%}    &   23.8\%    &   28.6\%    &   4.8\%   
    \\ 
    
    & It was easy to use the tool.    &   0\%   &   23.8\%    &   \textbf{33.3\%}    &   28.6\%    &   14.3\%   
    \\ 
    
    
    & I would likely use the tool to aid authoring again.   &   4.8\% &   9.5\% &   14.3\%    &   \textbf{47.6\%}    &   23.8\%   
    \\ 
    
    & The suggestions were useful for the authoring experience.   &   14.3\%    &   19\%  &   23.8\%    &   \textbf{28.6\%}    &   14.3\%   
    \\ 
    
    & The suggestions improved my authoring experience. &   23.8\%    &   4.8\% &   \textbf{33.3\%}    &   28.6\%    &   9.5\%   
    \\ 
    
    & How did you find the quality of the suggestions?   &   14.3\%    &   9.5\% &   28.6\%    &   \textbf{42.9\%}    &   4.8\%   
    \\ 

    & I think the suggestions were helpful in the creative process.  &   4.8\% &   4.8\% &   38.1\%    &   \textbf{42.9\%}    &   9.5\%   
    \\ 
    \midrule 

    \multirow{10}{*}{\rotatebox{90}{\textbf{System Usability Study}}} & I think that I would like to use this system frequently.     &   9.5\%    &   19.0\%  &   23.8\%  &   \textbf{28.6\%} &   19.0\%        \\ 
    
    & I found the system unnecessary complex &   23.8\% &   \textbf{47.6\%}    &   9.5\%    &   14.3\%    &   4.8\%    \\ 
    
    & I thought the system was easy to use    &   9.5\%   &   19.0\%    &   \textbf{28.6\%}    &   \textbf{28.6\%}    &  14.3\%   \\ 
    
    & 
    I would need the support of a technical person to be able to use this system.
  &   19.0\% &   19.0\%    &   \textbf{23.7\%}    &   14.3\%    &   \textbf{23.9\%}    \\ 
  
    & I found the various functions in this system were well integrated.   &   0.0\% &   9.5\% &   \textbf{42.9\%}    &   \textbf{42.9\%}    &   4.8\%   \\ 
    
    & I thought there was too much inconsistency in this system.  &   14.3\%    &   \textbf{52.4\%}  &   28.6\%    &   4.8\%    &   0.0\%    \\ 
    
    & I would imagine that most people would learn to use this system very quickly.   &   14.3\%    &   4.8\% &   23.8\%    &   \textbf{33.3\%}    &   23.8\%    \\ 
    
    & I found the system very cumbersome to use.    &   14.3\%    &   \textbf{33.3\%} &   \textbf{33.3\%}    &   19.0\%    &   0.0\%   \\ 
    
    & I felt very confident using the system.  &   9.5\% &   23.8\% &   19.0\%    &   \textbf{42.9\%}    &   4.8\%   \\ 
    
    & I needed to learn a lot of things before I could get going with this system.  &   14.3\% &   \textbf{33.3\%} &   9.5\%    &   28.6\%    &   14.3\%   \\ 
    
    \bottomrule \\
    
    \end{tabular}
    \label{table-user-study}                            

\end{table*}

\noindent \textbf{Results.} The tool identified all of the intended states, however it identified $10$ extra states as well. Of the $80$ affordance pre/post-conditions in the gold standard, $69$ were correctly identified, with an accuracy of $86.25\%$. Out of the $11$ cases that did not match, no condition was identified in one case and in the remaining $10$ cases, a false condition was identified. Some conditions were especially prone to errors. For example, out of $10$ mismatches, ``(Max, has, money)'' is responsible for $3$ mismatches and ``(Max, focus, typing)'' is responsible for $4$ mismatches. This occurs as in these cases word embeddings are not able to capture the semantics of the state. Also some of the states have a higher number of neighboring states in the embedding space. For example the state ``(Max, focus, typing)'' was confused $2$ times with ``(Max, Focus,  help\_customers)'' and also $2$ times with ``(Max, focus, play)''.
This illustrates the challenges associated with using state embeddings, with confused states having significant semantic resemblance, e.g., both are about ``focusing''. In future, we plan to address these issues by exploring other representations for states. Additionally, the system sometimes had difficulty handing complex sentences containing multiple verbs, where the sentence simplification module may have fed incorrect sentences to the state extraction module.

\subsection{User Study}

We recruited $21$ users (11 male, 10 female), between 21 and 49 years of age to evaluate our tool. The subjects were given a short tutorial describing the functions of the tool and asked to author their own character domain. After completing the task, they were asked two sets of questions: (1) tool-specific questions, (2) System Usability Study (SUS) questionnaire~\cite{brooke1996sus}. Responses were using a 5 point Likert scale \cite{likert1932technique} (Strongly Disagree, Disagree, Neutral, Agree, Strongly Agree). The SUS score was $61.2$ which is slightly below the average value of $68$. 
The questions and user responses are reported in Table~\ref{table-user-study}.
81\% of the users found the GUI of the tool to be convenient and easy to use.

\section{Conclusion}\label{sec:futureWork}

This paper presents a platform that allows users to author planning domains for intelligent virtual agents using a simple, intuitive natural language interface. We additionally explored the potential for using common sense knowledge to further inspire the creativity of the authors by providing suggestions about the domain. We quantitatively evaluate our tool by evaluating its output on a gold standard dataset. Two user studies are also conducted which highlight the usability and value of the natural language interface.
Our platform is currently integrated with a specific agent architecture. However, it can be generalized to universal planning domain definition languages (e.g., PDDL) in a straightforward manner. The ideas can even be further extended to non affordance based social interactive agent architectures as long as these have some consistent representation for their inner workings.
The natural understanding component of the proposed system is not perfect and it may face problems in the cases where an input sentence is very long and complex or it describes abstract concept which may be difficult to map to a specific rule. We plan to address these issues in the future. 

In our approach sentences were processed individually without taking into account the discourse information that links the states implied in the text. In the future, we would like to consider leveraging discourse information by considering the sequence of states/actions which are described in text. There has been work in the area of modeling event chains and scripts \cite{modi2014inducing,modi2016event,TACL968} and we would like to explore this line of research.



\section*{Acknowledgments}
Mubbasir Kapadia has been funded in part by NSF IIS-1703883 and NSF S\&AS-1723869.


\bibliographystyle{ACM-Reference-Format}  
\balance  
\bibliography{sample-aamas19}  


\begin{thebibliography}{00}


\ifx \showCODEN    \undefined \def \showCODEN     #1{\unskip}     \fi
\ifx \showDOI      \undefined \def \showDOI       #1{#1}\fi
\ifx \showISBNx    \undefined \def \showISBNx     #1{\unskip}     \fi
\ifx \showISBNxiii \undefined \def \showISBNxiii  #1{\unskip}     \fi
\ifx \showISSN     \undefined \def \showISSN      #1{\unskip}     \fi
\ifx \showLCCN     \undefined \def \showLCCN      #1{\unskip}     \fi
\ifx \shownote     \undefined \def \shownote      #1{#1}          \fi
\ifx \showarticletitle \undefined \def \showarticletitle #1{#1}   \fi
\ifx \showURL      \undefined \def \showURL       {\relax}        \fi
\providecommand\bibfield[2]{#2}
\providecommand\bibinfo[2]{#2}
\providecommand\natexlab[1]{#1}
\providecommand\showeprint[2][]{arXiv:#2}

\bibitem[\protect\citeauthoryear{Bindiganavale, Schuler, Allbeck, Badler,
  Joshi, and Palmer}{Bindiganavale et~al\mbox{.}}{2000}]%
        {bindiganavale2000dynamically}
\bibfield{author}{\bibinfo{person}{Rama Bindiganavale},
  \bibinfo{person}{William Schuler}, \bibinfo{person}{Jan~M Allbeck},
  \bibinfo{person}{Norman~I Badler}, \bibinfo{person}{Aravind~K Joshi}, {and}
  \bibinfo{person}{Martha Palmer}.} \bibinfo{year}{2000}\natexlab{}.
\newblock \showarticletitle{Dynamically Altering Agent Behaviors Using Natural
  Language Instructions}. In \bibinfo{booktitle}{{\em Departmental Papers
  (CIS)}}.
\newblock


\bibitem[\protect\citeauthoryear{Branavan, Kushman, Lei, and Barzilay}{Branavan
  et~al\mbox{.}}{2012}]%
        {branavan2012learning}
\bibfield{author}{\bibinfo{person}{SRK Branavan}, \bibinfo{person}{Nate
  Kushman}, \bibinfo{person}{Tao Lei}, {and} \bibinfo{person}{Regina
  Barzilay}.} \bibinfo{year}{2012}\natexlab{}.
\newblock \showarticletitle{Learning High-Level Planning from Text}. In
  \bibinfo{booktitle}{{\em Annual Meeting of the Association for Computational
  Linguistics (ACL)}}.
\newblock


\bibitem[\protect\citeauthoryear{Brooke et~al\mbox{.}}{Brooke
  et~al\mbox{.}}{1996}]%
        {brooke1996sus}
\bibfield{author}{\bibinfo{person}{John Brooke} {et~al\mbox{.}}}
  \bibinfo{year}{1996}\natexlab{}.
\newblock \showarticletitle{SUS-A Quick and Dirty Usability Scale}.
\newblock \bibinfo{journal}{{\em Usability Evaluation in Industry\/}}
  (\bibinfo{year}{1996}).
\newblock


\bibitem[\protect\citeauthoryear{Cassell, Sullivan, Churchill, and
  Prevost}{Cassell et~al\mbox{.}}{2000}]%
        {cassell2000embodied}
\bibfield{author}{\bibinfo{person}{Justine Cassell}, \bibinfo{person}{Joseph
  Sullivan}, \bibinfo{person}{Elizabeth Churchill}, {and}
  \bibinfo{person}{Scott Prevost}.} \bibinfo{year}{2000}\natexlab{}.
\newblock \bibinfo{booktitle}{{\em Embodied Conversational Agents}}.
\newblock \bibinfo{publisher}{MIT press}.
\newblock


\bibitem[\protect\citeauthoryear{Diakidoy, Kakas, Michael, and Miller}{Diakidoy
  et~al\mbox{.}}{2015}]%
        {diakidoy2015star}
\bibfield{author}{\bibinfo{person}{Irene-Anna Diakidoy},
  \bibinfo{person}{Antonis Kakas}, \bibinfo{person}{Loizos Michael}, {and}
  \bibinfo{person}{Rob Miller}.} \bibinfo{year}{2015}\natexlab{}.
\newblock \showarticletitle{STAR: A System of Argumentation for Story
  Comprehension and Beyond}. In \bibinfo{booktitle}{{\em AAAI Spring Symposium
  Series}}.
\newblock


\bibitem[\protect\citeauthoryear{Flatt and PLT}{Flatt and PLT}{2010}]%
        {plt-tr1}
\bibfield{author}{\bibinfo{person}{Matthew Flatt} {and} \bibinfo{person}{PLT}.}
  \bibinfo{year}{2010}\natexlab{}.
\newblock \bibinfo{booktitle}{{\em Reference: Racket}}.
\newblock \bibinfo{type}{{T}echnical {R}eport} PLT-TR-2010-1.
  \bibinfo{institution}{PLT Design Inc.}
\newblock


\bibitem[\protect\citeauthoryear{Fong, Nourbakhsh, and Dautenhahn}{Fong
  et~al\mbox{.}}{2003}]%
        {fong2003survey}
\bibfield{author}{\bibinfo{person}{Terrence Fong}, \bibinfo{person}{Illah
  Nourbakhsh}, {and} \bibinfo{person}{Kerstin Dautenhahn}.}
  \bibinfo{year}{2003}\natexlab{}.
\newblock \showarticletitle{A Survey of Socially Interactive Robots}.
\newblock \bibinfo{journal}{{\em Robotics and Autonomous Systems\/}}
  (\bibinfo{year}{2003}).
\newblock


\bibitem[\protect\citeauthoryear{Gebhard}{Gebhard}{2005}]%
        {gebhard2005alma}
\bibfield{author}{\bibinfo{person}{Patrick Gebhard}.}
  \bibinfo{year}{2005}\natexlab{}.
\newblock \showarticletitle{A{LMA}: a Layered Model of Affect}. In
  \bibinfo{booktitle}{{\em International Conference on Autonomous Agents and
  Multiagent Systems (AAMAS)}}.
\newblock


\bibitem[\protect\citeauthoryear{Gemignani, Bastianelli, and Nardi}{Gemignani
  et~al\mbox{.}}{2015}]%
        {gemignani2015teaching}
\bibfield{author}{\bibinfo{person}{Guglielmo Gemignani},
  \bibinfo{person}{Emanuele Bastianelli}, {and} \bibinfo{person}{Daniele
  Nardi}.} \bibinfo{year}{2015}\natexlab{}.
\newblock \showarticletitle{Teaching Robots Parametrized Executable Plans
  through Spoken Interaction}. In \bibinfo{booktitle}{{\em International
  Conference on Autonomous Agents and Multiagent Systems (AAMAS)}}.
\newblock


\bibitem[\protect\citeauthoryear{Gibson}{Gibson}{2014}]%
        {gibson2014ecological}
\bibfield{author}{\bibinfo{person}{James~J Gibson}.}
  \bibinfo{year}{2014}\natexlab{}.
\newblock \bibinfo{booktitle}{{\em The Ecological Approach to Visual
  Perception: Classic Edition}}.
\newblock \bibinfo{publisher}{Psychology Press}.
\newblock


\bibitem[\protect\citeauthoryear{Goldwasser and Roth}{Goldwasser and
  Roth}{2014}]%
        {goldwasser2014learning}
\bibfield{author}{\bibinfo{person}{Dan Goldwasser} {and} \bibinfo{person}{Dan
  Roth}.} \bibinfo{year}{2014}\natexlab{}.
\newblock \showarticletitle{Learning from Natural Instructions}.
\newblock \bibinfo{journal}{{\em Machine learning\/}} (\bibinfo{year}{2014}).
\newblock


\bibitem[\protect\citeauthoryear{Gratch and Marsella}{Gratch and
  Marsella}{2004}]%
        {gratch2004domain}
\bibfield{author}{\bibinfo{person}{Jonathan Gratch} {and}
  \bibinfo{person}{Stacy Marsella}.} \bibinfo{year}{2004}\natexlab{}.
\newblock \showarticletitle{A Domain-Independent Framework for Modeling
  Emotion}.
\newblock \bibinfo{journal}{{\em Cognitive Systems Research\/}}
  (\bibinfo{year}{2004}).
\newblock


\bibitem[\protect\citeauthoryear{Harrington and O`Connell}{Harrington and
  O`Connell}{2016}]%
        {harrington2016video}
\bibfield{author}{\bibinfo{person}{Brian Harrington} {and}
  \bibinfo{person}{Michael O`Connell}.} \bibinfo{year}{2016}\natexlab{}.
\newblock \showarticletitle{Video Games as Virtual Teachers: Prosocial Video
  Game Use by Children and Adolescents from Different Socioeconomic Groups is
  Associated with Increased Empathy and Prosocial Behaviour}.
\newblock \bibinfo{journal}{{\em Computers in Human Behavior\/}}
  (\bibinfo{year}{2016}).
\newblock


\bibitem[\protect\citeauthoryear{Harris}{Harris}{1954}]%
        {harris1954distributional}
\bibfield{author}{\bibinfo{person}{Zellig~S Harris}.}
  \bibinfo{year}{1954}\natexlab{}.
\newblock \bibinfo{booktitle}{{\em Distributional Structure}}.
\newblock \bibinfo{publisher}{Englewood Cliffs, NJ: Prentice-Hall}.
\newblock


\bibitem[\protect\citeauthoryear{Hayton, Porteous, Ferreira, Lindsay, and
  Read}{Hayton et~al\mbox{.}}{2017}]%
        {hayton2017storyframer}
\bibfield{author}{\bibinfo{person}{Thomas Hayton}, \bibinfo{person}{Julie
  Porteous}, \bibinfo{person}{Joao~F Ferreira}, \bibinfo{person}{Alan Lindsay},
  {and} \bibinfo{person}{Jonathon Read}.} \bibinfo{year}{2017}\natexlab{}.
\newblock \showarticletitle{StoryFramer: From Input Stories to Output Planning
  Models}.
\newblock


\bibitem[\protect\citeauthoryear{Honnibal and Johnson}{Honnibal and
  Johnson}{2015}]%
        {spacy-parser}
\bibfield{author}{\bibinfo{person}{Matthew Honnibal} {and}
  \bibinfo{person}{Mark Johnson}.} \bibinfo{year}{2015}\natexlab{}.
\newblock \showarticletitle{An Improved Non-monotonic Transition System for
  Dependency Parsing}. In \bibinfo{booktitle}{{\em Conference on Empirical
  Methods in Natural Language Processing (EMNLP)}}.
\newblock


\bibitem[\protect\citeauthoryear{Honnibal and Montani}{Honnibal and
  Montani}{2017}]%
        {spacy2}
\bibfield{author}{\bibinfo{person}{Matthew Honnibal} {and}
  \bibinfo{person}{Ines Montani}.} \bibinfo{year}{2017}\natexlab{}.
\newblock \showarticletitle{spaCy 2: Natural Language Understanding with Bloom
  Embeddings, Convolutional Neural Networks and Incremental Parsing}.
\newblock  (\bibinfo{year}{2017}).
\newblock


\bibitem[\protect\citeauthoryear{Kallmann and Thalmann}{Kallmann and
  Thalmann}{1999}]%
        {kallmann1999modeling}
\bibfield{author}{\bibinfo{person}{Marcelo Kallmann} {and}
  \bibinfo{person}{Daniel Thalmann}.} \bibinfo{year}{1999}\natexlab{}.
\newblock \showarticletitle{Modeling Objects for Interaction Tasks}.
\newblock In \bibinfo{booktitle}{{\em Computer Animation \& Simulation}}.
\newblock


\bibitem[\protect\citeauthoryear{Kamath and Das}{Kamath and Das}{2018}]%
        {SemanticParsing2018}
\bibfield{author}{\bibinfo{person}{Aishwarya Kamath} {and}
  \bibinfo{person}{Rajarshi Das}.} \bibinfo{year}{2018}\natexlab{}.
\newblock \showarticletitle{A Survey on Semantic Parsing}.
\newblock \bibinfo{journal}{{\em CoRR\/}}  \bibinfo{volume}{abs/1812.00978}
  (\bibinfo{year}{2018}).
\newblock


\bibitem[\protect\citeauthoryear{Kim and Baylor}{Kim and Baylor}{2008}]%
        {kim2008virtual}
\bibfield{author}{\bibinfo{person}{ChanMin Kim} {and} \bibinfo{person}{Amy~L
  Baylor}.} \bibinfo{year}{2008}\natexlab{}.
\newblock \showarticletitle{A Virtual Change Agent: Motivating Preservice
  Teachers to Integrate Technology in their Future Classrooms}.
\newblock \bibinfo{journal}{{\em Journal of Educational Technology \&
  Society\/}} (\bibinfo{year}{2008}).
\newblock


\bibitem[\protect\citeauthoryear{Kollar, Perera, Nardi, and Veloso}{Kollar
  et~al\mbox{.}}{2013}]%
        {kollar2013learning}
\bibfield{author}{\bibinfo{person}{Thomas Kollar}, \bibinfo{person}{Vittorio
  Perera}, \bibinfo{person}{Daniele Nardi}, {and} \bibinfo{person}{Manuela
  Veloso}.} \bibinfo{year}{2013}\natexlab{}.
\newblock \showarticletitle{Learning Environmental Knowledge from Task-Based
  Human-Robot Dialog}. In \bibinfo{booktitle}{{\em IEEE International
  Conference on Robotics and Automation (ICRA)}}.
\newblock


\bibitem[\protect\citeauthoryear{Likert}{Likert}{1932}]%
        {likert1932technique}
\bibfield{author}{\bibinfo{person}{Rensis Likert}.}
  \bibinfo{year}{1932}\natexlab{}.
\newblock \showarticletitle{A Technique for the Measurement of Attitudes}.
\newblock \bibinfo{journal}{{\em Archives of psychology\/}}
  (\bibinfo{year}{1932}).
\newblock


\bibitem[\protect\citeauthoryear{Lindsay, Read, Ferreira, Hayton, Porteous, and
  Gregory}{Lindsay et~al\mbox{.}}{2017}]%
        {lindsay2017framer}
\bibfield{author}{\bibinfo{person}{Alan Lindsay}, \bibinfo{person}{Jonathon
  Read}, \bibinfo{person}{Joao~F Ferreira}, \bibinfo{person}{Thomas Hayton},
  \bibinfo{person}{Julie Porteous}, {and} \bibinfo{person}{Peter~J Gregory}.}
  \bibinfo{year}{2017}\natexlab{}.
\newblock \showarticletitle{Framer: Planning Models From Natural Language
  Action Descriptions}. In \bibinfo{booktitle}{{\em International Conference on
  Automated Planning and Scheduling (ICAPS)}}.
\newblock


\bibitem[\protect\citeauthoryear{Matsuyama, Bhardwaj, Zhao, Romeo, Akoju, and
  Cassell}{Matsuyama et~al\mbox{.}}{2016}]%
        {matsuyama2016socially}
\bibfield{author}{\bibinfo{person}{Yoichi Matsuyama}, \bibinfo{person}{Arjun
  Bhardwaj}, \bibinfo{person}{Ran Zhao}, \bibinfo{person}{Oscar Romeo},
  \bibinfo{person}{Sushma Akoju}, {and} \bibinfo{person}{Justine Cassell}.}
  \bibinfo{year}{2016}\natexlab{}.
\newblock \showarticletitle{Socially-Aware Animated Intelligent Personal
  Assistant Agent}.
\newblock


\bibitem[\protect\citeauthoryear{Mc{D}ermott, Ghallab, Howe, Knoblock, Ram,
  Veloso, Weld, and Wilkins}{Mc{D}ermott et~al\mbox{.}}{1998}]%
        {mcdermott98pddl}
\bibfield{author}{\bibinfo{person}{Drew Mc{D}ermott}, \bibinfo{person}{Malik
  Ghallab}, \bibinfo{person}{Adele Howe}, \bibinfo{person}{Craig Knoblock},
  \bibinfo{person}{Ashwin Ram}, \bibinfo{person}{Manuela Veloso},
  \bibinfo{person}{Daniel Weld}, {and} \bibinfo{person}{David Wilkins}.}
  \bibinfo{year}{1998}\natexlab{}.
\newblock \bibinfo{booktitle}{{\em PDDL - The Planning Domain Definition
  Language}}.
\newblock \bibinfo{type}{{T}echnical {R}eport}.
\newblock


\bibitem[\protect\citeauthoryear{Mehrabian}{Mehrabian}{1996}]%
        {mehrabian1996pleasure}
\bibfield{author}{\bibinfo{person}{Albert Mehrabian}.}
  \bibinfo{year}{1996}\natexlab{}.
\newblock \showarticletitle{Pleasure-Arousal-Dominance: a General Framework for
  Describing and Measuring Individual Differences in Temperament}.
\newblock \bibinfo{journal}{{\em Current Psychology\/}} (\bibinfo{year}{1996}).
\newblock


\bibitem[\protect\citeauthoryear{Mikolov, Chen, Corrado, and Dean}{Mikolov
  et~al\mbox{.}}{2013a}]%
        {mikolov2013efficient}
\bibfield{author}{\bibinfo{person}{Tomas Mikolov}, \bibinfo{person}{Kai Chen},
  \bibinfo{person}{Greg Corrado}, {and} \bibinfo{person}{Jeffrey Dean}.}
  \bibinfo{year}{2013}\natexlab{a}.
\newblock \showarticletitle{Efficient Estimation of Word Representations in
  Vector Space}.
\newblock \bibinfo{journal}{{\em arXiv preprint arXiv:1301.3781\/}}
  (\bibinfo{year}{2013}).
\newblock


\bibitem[\protect\citeauthoryear{Mikolov, Sutskever, Chen, Corrado, and
  Dean}{Mikolov et~al\mbox{.}}{2013b}]%
        {mikolov2013distributed}
\bibfield{author}{\bibinfo{person}{Tomas Mikolov}, \bibinfo{person}{Ilya
  Sutskever}, \bibinfo{person}{Kai Chen}, \bibinfo{person}{Greg~S Corrado},
  {and} \bibinfo{person}{Jeff Dean}.} \bibinfo{year}{2013}\natexlab{b}.
\newblock \showarticletitle{Distributed Representations of Words and Phrases
  and their Compositionality}. In \bibinfo{booktitle}{{\em Annual Conference on
  Neural Information Processing Systems (NIPS)}}.
\newblock


\bibitem[\protect\citeauthoryear{Modi}{Modi}{2016}]%
        {modi2016event}
\bibfield{author}{\bibinfo{person}{Ashutosh Modi}.}
  \bibinfo{year}{2016}\natexlab{}.
\newblock \showarticletitle{Event Embeddings for Semantic Script Modeling}. In
  \bibinfo{booktitle}{{\em Proceedings of The 20th SIGNLL Conference on
  Computational Natural Language Learning}}.
\newblock


\bibitem[\protect\citeauthoryear{Modi and Titov}{Modi and Titov}{2014}]%
        {modi2014inducing}
\bibfield{author}{\bibinfo{person}{Ashutosh Modi} {and} \bibinfo{person}{Ivan
  Titov}.} \bibinfo{year}{2014}\natexlab{}.
\newblock \showarticletitle{Inducing Neural Models of Script Knowledge}. In
  \bibinfo{booktitle}{{\em Conference on Computational Natural Language
  Learning (CoNLL)}}.
\newblock


\bibitem[\protect\citeauthoryear{Modi, Titov, Demberg, Sayeed, and Pinkal}{Modi
  et~al\mbox{.}}{2017}]%
        {TACL968}
\bibfield{author}{\bibinfo{person}{Ashutosh Modi}, \bibinfo{person}{Ivan
  Titov}, \bibinfo{person}{Vera Demberg}, \bibinfo{person}{Asad Sayeed}, {and}
  \bibinfo{person}{Manfred Pinkal}.} \bibinfo{year}{2017}\natexlab{}.
\newblock \showarticletitle{Modelling Semantic Expectation: Using Script
  Knowledge for Referent Prediction}.
\newblock \bibinfo{journal}{{\em Transactions of the Association for
  Computational Linguistics\/}}  \bibinfo{volume}{5} (\bibinfo{year}{2017}),
  \bibinfo{pages}{31--44}.
\newblock
\showISSN{2307-387X}
\showURL{%
\url{https://transacl.org/ojs/index.php/tacl/article/view/968}}


\bibitem[\protect\citeauthoryear{Pecune, Ochs, Marsella, and Pelachaud}{Pecune
  et~al\mbox{.}}{2016}]%
        {pecune2016socrates}
\bibfield{author}{\bibinfo{person}{Florian Pecune}, \bibinfo{person}{Magalie
  Ochs}, \bibinfo{person}{Stacy Marsella}, {and} \bibinfo{person}{Catherine
  Pelachaud}.} \bibinfo{year}{2016}\natexlab{}.
\newblock \showarticletitle{Socrates: from Social Relation to Attitude
  Expressions}. In \bibinfo{booktitle}{{\em International Conference on
  Autonomous Agents \& Multiagent Systems (AAMAS)}}.
\newblock


\bibitem[\protect\citeauthoryear{Perera and Veloso}{Perera and Veloso}{2014}]%
        {perera2014task}
\bibfield{author}{\bibinfo{person}{Vittorio Perera} {and}
  \bibinfo{person}{Manuela Veloso}.} \bibinfo{year}{2014}\natexlab{}.
\newblock \showarticletitle{Task Based Dialog for Service Mobile Robot}. In
  \bibinfo{booktitle}{{\em AAAI Fall Symposium Series}}.
\newblock


\bibitem[\protect\citeauthoryear{Pomarlan and Bateman}{Pomarlan and
  Bateman}{2018}]%
        {Pomarlan-2018}
\bibfield{author}{\bibinfo{person}{Mihai Pomarlan} {and} \bibinfo{person}{John
  Bateman}.} \bibinfo{year}{2018}\natexlab{}.
\newblock \showarticletitle{Robot Program Construction via Grounded Natural
  Language Semantics \& Simulation}. In \bibinfo{booktitle}{{\em International
  Conference on Autonomous Agents and MultiAgent Systems (AAMAS)}}.
\newblock


\bibitem[\protect\citeauthoryear{Rehurek and Sojka}{Rehurek and Sojka}{2010}]%
        {rehurek_lrec}
\bibfield{author}{\bibinfo{person}{Radim Rehurek} {and} \bibinfo{person}{Petr
  Sojka}.} \bibinfo{year}{2010}\natexlab{}.
\newblock \showarticletitle{Software Framework for Topic Modelling with Large
  Corpora}. In \bibinfo{booktitle}{{\em {Proceedings of the LREC Workshop on
  New Challenges for NLP Frameworks}}}.
\newblock


\bibitem[\protect\citeauthoryear{Reiss and Havercamp}{Reiss and
  Havercamp}{1998}]%
        {reiss1998toward}
\bibfield{author}{\bibinfo{person}{Steven Reiss} {and} \bibinfo{person}{Susan~M
  Havercamp}.} \bibinfo{year}{1998}\natexlab{}.
\newblock \showarticletitle{Toward a Comprehensive Assessment of Fundamental
  Motivation: Factor Structure of the Reiss Profiles}.
\newblock \bibinfo{journal}{{\em Psychological Assessment\/}}
  (\bibinfo{year}{1998}).
\newblock


\bibitem[\protect\citeauthoryear{Ribeiro, Pereira, Di~Tullio, Alves-Oliveira,
  and Paiva}{Ribeiro et~al\mbox{.}}{2014}]%
        {ribeiro2014thalamus}
\bibfield{author}{\bibinfo{person}{Tiago Ribeiro}, \bibinfo{person}{Andr{\'e}
  Pereira}, \bibinfo{person}{Eugenio Di~Tullio}, \bibinfo{person}{Patr{\i}cia
  Alves-Oliveira}, {and} \bibinfo{person}{Ana Paiva}.}
  \bibinfo{year}{2014}\natexlab{}.
\newblock \showarticletitle{From Thalamus to Skene: High-level Behaviour
  Planning and Managing for Mixed-Reality Characters}. In
  \bibinfo{booktitle}{{\em Proceedings of the IVA 2014 Workshop on
  Architectures and Standards for IVAs}}.
\newblock


\bibitem[\protect\citeauthoryear{Saggion}{Saggion}{2017}]%
        {saggion2017automatic}
\bibfield{author}{\bibinfo{person}{Horacio Saggion}.}
  \bibinfo{year}{2017}\natexlab{}.
\newblock \showarticletitle{Automatic Text Simplification}.
\newblock \bibinfo{journal}{{\em Synthesis Lectures on Human Language
  Technologies\/}} (\bibinfo{year}{2017}).
\newblock


\bibitem[\protect\citeauthoryear{Sanghrajka, Schriber, Gross, and
  Kapadia}{Sanghrajka et~al\mbox{.}}{2018}]%
        {sanghrajka2018computer}
\bibfield{author}{\bibinfo{person}{Rushit Sanghrajka}, \bibinfo{person}{Sasha
  Schriber}, \bibinfo{person}{Markus~H Gross}, {and} \bibinfo{person}{Mubbasir
  Kapadia}.} \bibinfo{year}{2018}\natexlab{}.
\newblock \showarticletitle{Computer-Assisted Authoring for Natural Language
  Story Scripts}. In \bibinfo{booktitle}{{\em AAAI Conference on Innovative
  Applications of Artificial Intelligence (IAAI)}}.
\newblock


\bibitem[\protect\citeauthoryear{Sil and Yates}{Sil and Yates}{2011}]%
        {sil2011extracting}
\bibfield{author}{\bibinfo{person}{Avirup Sil} {and} \bibinfo{person}{Alexander
  Yates}.} \bibinfo{year}{2011}\natexlab{}.
\newblock \showarticletitle{Extracting STRIPS Representations of Actions and
  Events}. In \bibinfo{booktitle}{{\em International Conference Recent Advances
  in Natural Language Processing (RANLP)}}.
\newblock


\bibitem[\protect\citeauthoryear{Speer and Havasi}{Speer and Havasi}{2012}]%
        {speer2012representing}
\bibfield{author}{\bibinfo{person}{Robert Speer} {and}
  \bibinfo{person}{Catherine Havasi}.} \bibinfo{year}{2012}\natexlab{}.
\newblock \showarticletitle{Representing General Relational Knowledge in
  ConceptNet 5}.
\newblock


\bibitem[\protect\citeauthoryear{Suchanek, Kasneci, and Weikum}{Suchanek
  et~al\mbox{.}}{2007}]%
        {suchanek2007yago}
\bibfield{author}{\bibinfo{person}{Fabian~M Suchanek}, \bibinfo{person}{Gjergji
  Kasneci}, {and} \bibinfo{person}{Gerhard Weikum}.}
  \bibinfo{year}{2007}\natexlab{}.
\newblock \showarticletitle{Yago: a Core of Semantic Knowledge}. In
  \bibinfo{booktitle}{{\em International Conference on World Wide Web}}.
\newblock


\bibitem[\protect\citeauthoryear{Summerfield}{Summerfield}{2007}]%
        {summerfield2007rapid}
\bibfield{author}{\bibinfo{person}{Mark Summerfield}.}
  \bibinfo{year}{2007}\natexlab{}.
\newblock \bibinfo{booktitle}{{\em Rapid GUI Programming with Python and Qt:
  The Definitive Guide to PyQt Programming (paperback)}}.
\newblock \bibinfo{publisher}{Pearson Education}.
\newblock


\bibitem[\protect\citeauthoryear{Theune, Faas, Nijholt, and Heylen}{Theune
  et~al\mbox{.}}{2003}]%
        {theune2003virtual}
\bibfield{author}{\bibinfo{person}{Mari{\"e}t Theune}, \bibinfo{person}{Sander
  Faas}, \bibinfo{person}{Anton Nijholt}, {and} \bibinfo{person}{Dirk Heylen}.}
  \bibinfo{year}{2003}\natexlab{}.
\newblock \showarticletitle{The Virtual Storyteller: Story Creation by
  Intelligent Agents}. In \bibinfo{booktitle}{{\em Proceedings of the
  Technologies for Interactive Digital Storytelling and Entertainment
  Conference (TIDSE)}}.
\newblock


\bibitem[\protect\citeauthoryear{Yordanova}{Yordanova}{2017}]%
        {yordanova2017texttohbm}
\bibfield{author}{\bibinfo{person}{Kristina Yordanova}.}
  \bibinfo{year}{2017}\natexlab{}.
\newblock \showarticletitle{TextToHBM: A Generalised Approach to Learning
  Models of Human Behaviour for Activity Recognition from Textual
  Instructions}. In \bibinfo{booktitle}{{\em Workshops at the Thirty-First AAAI
  Conference on Artificial Intelligence}}.
\newblock


\bibitem[\protect\citeauthoryear{Youssef, Chollet, Jones, Sabouret, Pelachaud,
  and Ochs}{Youssef et~al\mbox{.}}{2015}]%
        {youssef2015towards}
\bibfield{author}{\bibinfo{person}{Atef~Ben Youssef}, \bibinfo{person}{Mathieu
  Chollet}, \bibinfo{person}{Haza{\"e}l Jones}, \bibinfo{person}{Nicolas
  Sabouret}, \bibinfo{person}{Catherine Pelachaud}, {and}
  \bibinfo{person}{Magalie Ochs}.} \bibinfo{year}{2015}\natexlab{}.
\newblock \showarticletitle{Towards a Socially Adaptive Virtual Agent}. In
  \bibinfo{booktitle}{{\em International Conference on Intelligent Virtual
  Agents (IVA)}}.
\newblock


\end{thebibliography}

\end{document}